# Prediction of User Request and Complaint in Spoken Customer-Agent Conversations

*Nikola Lackovic[1, 2], Claude Montacié[2], Gauthier Lalande[1] and Marie-José Caraty[2]*

[1]Malakof Humanis, 21 rue Laffitte, Paris, 75009, France
[2]STIH Laboratory, Sorbonne University, 28 rue Serpente, Paris, 75006, France
nikola.lackovic@malakoffhumanis.com, claude.montacie@sorbonne-universite.fr,
gauthier.lalande@malakoffhumanis.com, marie-jose.caraty@u-paris.fr

## Abstract

We present the corpus called HealthCall. This was recorded in real-life conditions in the call center of Malakof Humanis. It includes two separate audio channels, the first one for the customer and the second one for the agent. Each conversation was anonymized respecting the General Data Protection Regulation. This corpus includes a transcription of the spoken conversations and was divided into two sets: Train and Devel sets. Two important customer relationship management tasks were assessed on the HealthCall corpus: Automatic prediction of type of user requests and complaints detection. For this purpose, we have investigated 14 feature sets: 6 linguistic feature sets, 6 audio feature sets and 2 vocal interaction feature sets. We have used Bidirectional Encoder Representation from Transformers models for the linguistic features, openSMILE and Wav2Vec 2.0 for the audio features. The vocal interaction feature sets were designed and developed from Turn Takings. The results show that the linguistic features always give the best results (91.2% for the Request task and 70.3% for the Complaint task). The Wav2Vec 2.0 features seem more suitable for these two tasks than the ComPaRe16 features. Vocal interaction features outperformed ComPaRe16 features on Complaint task with a 57% rate achieved with only six features.

**Index Terms**: call center corpus, customer relationship management, linguistic features, audio features, vocal interaction features, request, complaint.

## 1. Introduction

Customer Relationship Management (CRM) is an important framework for managing the organization's interactions with its customers [1]. The organization interacts with its customer most often through a Call Center (CC) [2]. It allows customers to express their requests via communication channels such as phone, email or online chat. Using speech analytics [3, 4] to model, characterize and identify these phone conversations has become an indispensable tool for CRM. It is also an important field of application for speech processing and more particularly for the study of vocal interactions.

Speech analytics has already been used in several areas of customer relationship such as monitoring the quality of interactions [5, 6, 7], user satisfaction analysis [8, 9, 10, 11], classifying the type of customer requests [12, 13, 14] and detecting complaints [15, 16].

Monitoring the quality of interactions in call centers aims at measuring the performance of a customer service. In the related works, quality measures were computed in combining a transcript from an ASR (Automatic Speech Recognition) system, detection of keywords or keyphrases, number and type of hesitations, average silence durations [5], text similarity measures [6] and audio features [7]. User satisfaction is normally assessed after the customer-agent interactions with its customers through questionnaires or oral or written interviews. In order to predict self-reported satisfaction in spoken conversations, various studies have investigated the relationship between satisfaction and emotions [8, 11]. Various prosodic, linguistic [9] and vocal interaction features [8, 10] have also been studied. Automatic conversation classification in call centers can provide real-time feedback on the evolution of the type of customer requests. Most of the work is based on an analysis of conversation transcripts [12]. The goal is to develop methods that are robust to errors in current ASR systems [13] and take into account conversational features such as turn taking [14]. Customer complaints have been an important feedback for modern companies to improve customer loyalty. The characteristics of telephone complaints were studied using the usual emotional characteristics [15] such as those related to anger [16].

For this study, based on related work, we were interested in the respective contributions of various features of spoken conversation (vocal interaction, audio and linguistics) in two tasks related to CRM: the classification of customer request types and the detection of customer complaints. The article was organized as follows: The call center database and its training and development sets are presented in Section 2. The implementation of the textual data modeling corresponding to the spoken conversations is presented in Section 3. Vocal interaction features computed from customer-agent dialogue are described in Section 4. In Section 5, two types of audio features sets are presented. The experiments are described in Section 6. The last Section concludes the study.

## 2. The HealthCall Corpus

Several corpora have been collected in call centers in the past and used for conversation analysis. We can mention CU Call Center [12], DECODA [17], AlloSat[18] and ClovaCall[19] corpora. The HealthCall (HC) corpus we provide [20] is based on real audio interactions between call center agents and customers. These customers call the Malakof Humanis (MH) health insurance company to resolve a problem or request information. This corpus was designed to study natural spoken conversations and to predict CRM annotations made by human agents from vocal interactions. This corpus is composed of 2,416 spoken conversations of varying duration (from a few minutes to several tens of minutes). Each conversation is anonymized respecting the General Data Protection Regulation (GDPR) [21] recommendation and includes in addition to CRM annotations a transcription made by Allomedia company.

## 2.1. Speech Analytics Process

We have implemented a Speech Analytics (SA) process [3, 4] which made it possible to collect the HC-corpus. The SA-process provided call center agents with indicators to annotate conversations in real time.

The speakers are customers or MH agents. The speakers can be female or male. The annotators were trained to address the problems of multi-class labelling. For the HC-corpus, we selected two classes, each with two choices, from the CRM annotations: Request and Complaint.

The Request class corresponds to the type of client requests. Two labels representing a majority of the applications were chosen: process and member. The first label called process corresponds on the healthcare processing. This is to address an issue regarding the reimbursement of a healthcare process. The second label is called member. It is a question of answering the problems of adhesion or subscription to the company.

The Complaint class is based on the presence or absence in the CRM annotations of a customer complaint. The labels are yes or not.

## 2.2. Statistics and Qualitative Properties

The spoken conversations in the HC corpus were recorded in the first half of 2021. We present on the table 1 several characteristics of the corpus.

Table 1: *Corpus **Statistics***

| Corpus Properties | Values |
|---|---|
| **General Description** | |
| Number of conversations | 2,416 |
| Total Duration | 251 hours 53 min |
| Max Duration | 46 min 36 sec |
| Min Duration | 1 min 18 sec |
| **Experimental Set** | |
| Conversations Train | 1,214 |
| Request (Process) Train | 650 |
| Request (Member) Train | 564 |
| Complaint Train | 461 |
| No Complaint Train | 753 |
| | |
| Conversations Dev | 1,202 |
| Request (Process) Dev | 584 |
| Request (Member) Dev | 618 |
| Complaint Dev | 455 |
| No Complaint Dev | 747 |

The corpus is in French and the audio was sampled at 8 khz rate and encoded at a rate of 32 kb/s. Each conversation was recorded on two separate audio channels, the first one for the customer and the second one for the agent. The corpus was divided into two sets: Train and Devel sets. There is no common speaker (customer and agent) for these two sets. The transcription files are in JSON format, several tokens are anonymized for data privacy purposes, such as name, surname, Bank account number, etc. The list of anonymization tokens can be found in the repository [20].

## 2.3. Data Gathering and Annotation Process

We have integrated several heterogenous company datasets in order to retrieve the information of agent annotations related to spoken conversations and their metadata.

The annotation process is at the end of an interactive call between a customer and an agent. It is described as follows:

- The customer calls the corporate call line and accesses a vocal server system, -in case of no resolution of his problem, he is redirected to an agent.
- The conversation between the customer and the agent to solve the problem starts after a waiting time in a call server queue.
- After the conversation, the agent annotates it with CRM tags and adds a comment on the feeling of the conversation.

## 2.4. Anonymization Process

In order to preserve the privacy of customers and agents, the speech segments corresponding to personal data are replaced with anonymized sounds. We did not choose silence or white noise for the anonymized sound, but an analysis-synthesis of the original signal by stylizing the formantic curve while keeping the prosodic information [22]. To avoid the recognition of the identity of clients or agents by speaker recognition techniques, the same analysis-synthesis techniques provided by the *praat* software [23] have been used by slightly but randomly modifying the formants on the whole speech signal.

# 3. Linguistic features

Lexical cues corresponding to an audio signal can be used to discriminate the type of audio segment (request or complaint) from methods such as Continuous Bag-Of-Words (CBOW) [24], Pointwise Mutual Information (PMI) [25], Term Frequency-Inverse Document Frequency (TFIDF), or more recently by a Bidirectional Encoder Representation from Transformers (BERT) model [26].

We have used a French version of the BERT model called Camembert [27] available at [28]. Our goal is to represent the text corresponding to a conversation by a single vector of linguistic features. We have chosen to use the Camembert-based integration layer which allows to extract a 768-dimensional feature vector. Other solutions were possible such as the weighted sum of several layers [29].

In this model, the size of the encoded text is limited to a sequence of 512 tokens which is insufficient in a case where conversations can exceed 5,000 words. We have chosen to truncate long conversations to 512 tokens. Two usual truncation methods [30] were chosen: the Head method which consists in keeping only the first 512 tokens and the Tail method which models only the last 512 tokens. Three types of textual data have been modeled by these two methods: the customer textual data, the agent textual data and the whole textual data. This was done using the standard text encoder and tokenizer from the pytorch library [31], and feature extraction was then performed on these tokenized textual data. The BERT model was not tuned on the Train data to allow reproducibility of the experiments.

At last, we obtained six sets of features of size 768 for each conversation: Hc (**H**ead truncation for **c**ustomer text data), Ha (**H**ead truncation for **a**gent text data), Hw (**H**ead truncation for the **w**hole text data), Tc (**T**ail truncation for **c**ustomer text data), Ta (**T**ail truncation for **a**gent text data), Tw (**T**ail truncation for the **w**hole text data).

## 4. Vocal interaction features

Vocal interaction features were usually computed from talkspurts (contiguous speech intervals, with short internal pauses) and silent periods. Conversation is a social interaction between two or more people, where taking turns to talk is naturally observed. Since it is difficult to speak and listen at the same time during a conversation, people must coordinate. Humans usually manage to fluent turns of speech with very small gaps and little overlap except in cases of misunderstanding or conflict. In the pioneering work of Sacks, Schegloff, and Jefferson [32], an organizational model of turn-taking for conversation has been introduced. In previous works [33, 34], a Markov model has been computed from the gaps (long silences), the switching pauses between different speakers, the simultaneous speech, and the single speaker vocalizations.

We have defined eight types of temporal segments to label conversations using the temporal codes associated with the ASR transcriptions: customer is speaking alone (S1), agent is speaking alone (S2), agent begins to speak while customer is speaking (S3), customer begins to speak while agent is speaking (S4), switching pauses between agent and customer (S5), switching pauses between customer and agent (S6), customer stops speaking then starts speaking again (S7), agent stops speaking then starts speaking again (S8). Figure 1 shows the possible transitions between the segment types.

For each type of segment St, we computed the following statistics on the duration of the segments: Minimum (Mint), Maximum (Maxt), Mean (Meant), Standard deviation (Sdt), Kurtosis (Kt) and Skewness (Skt). We also calculated the proportion of a segment St in duration (Tt) and number (Nt) to all the segments. This provides 64 vocal interaction features to describe a spoken conversation. We call this vocal Interaction feature set TT.

The relevance of all the features is given by the information gain [35] which is computed on the Train set with the following formula:

$$H(class) - H(class/feature) \qquad (1)$$

where Shannon entropy H is estimated from a table of contingency and class (Request or Complaint). Features for which the information gain is greater than zero are considered as relevant.

For the Complaint task, six features out of 64 are relevant and selected for the Complaint classifier. The ranking order of relevance is the following: T7, Max7, Sk5, K5, Mean7 and Mean5. We call this feature set TTc. We can notice that all these features correspond to statistics on the switching pauses between agent and customer and on the customer pauses.

No relevant feature was found for the request task. We kept the feature set TT for this task.

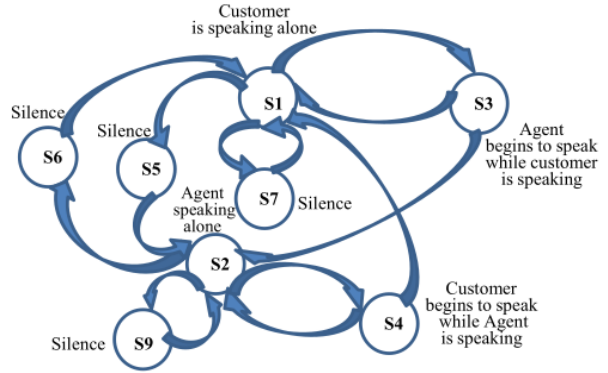

Figure 1: *Diagram of customer-agent dialogue*

## 5. Audio features

Our goal is to get a single audio feature vector for the entire conversation. It avoids the use of the expensive procedures of time warping between sequences of different lengths such as dynamic programming algorithms. We used two toolkits for that: openSMILE [36] and Wav2Vec 2.0 [37].

The OpenSMILE toolkit is widely used and has been shown to provide good performance in most of the many paralinguistic applications. There are several audio feature sets associated with this toolkit such as ComPare2016 set used for ComParE challenges. This audio feature set consists of 6,373 static features. It is based on several Low-Level Descriptors (LLDs) that are computed from short overlapping windows of the audio signal. These LLDs comprise spectral and prosodic coefficients such as band-energies, pitch, harmonics-to-noise ratio and jitter. On these LLDs, various functionals (global statistical functions) are computed to obtain feature vectors of equal size for each speech utterance. Some functionals aim at estimating the spatial variability (e.g., mean, standard deviation, quartiles 1-3) and others aim at the temporal variability (e.g., peaks, linear regression slope). At last, we obtained three sets of features of size 6,373 for each conversation: Cc (**C**omPare2016 features from **c**ustomer channel), Ca (**C**omPare2016 features from **a**gent channel) and Cj (**C**omPare2016 features from **j**oint customer-agent channel).

The Wav2Vec 2.0 (W2V2) toolkit is based on a convolutional neural network encoder, a contextualized transformer network and a quantization module. This approach is similar to the masked language modeling used in BERT. This neural network model was developed for phoneme recognition and has as input a speech signal. The W2V2 toolkit allow to extract from the speech signal a 768-dimensional feature vector every 20 ms for a given layer of the encoder. Each layer yields a different representation that might be more or less suitable for a task than a later or previous layer [37]. The W2V2 features have already been used in other speech-related applications, such as emotion recognition, accent identification and speech impairment detection. We use a pre-trained model on 960 hours of unlabeled speech from the LibriSpeech corpus [38]. We have chosen to extract the features from the penultimate layer. Four usual statistics functionals (Mean, Standard deviation, Kurtosis and Skewness) have been computed on the sequence of W2V2 features to obtain a single audio 4x768-dimensional feature vector for the entire conversation. The W2V2 model was not tuned on the Train data to allow reproducibility of the experiments. At last, we obtained three sets of features of size 3072 for each conversation: Wc (W2V2 features from customer channel), Wa (W2V2 features from agent channel) and Wj (W2V2 features from joint customer-agent channel)

# 6. Experimental Results

Support Vector Machines (SVM) classifier with linear Kernel was used for the classification of the types of requests and the detection of complaints. Scikit-learn toolbox [39] was used for the implementation of the classifier. Posterior probabilities were computed by the isotonic regression method [40]. Complexity parameters of the SVM classifier were optimized to maximize the Unweighted Average Recall (UAR) on the Devel set.

## 6.1. Linguistic features

Table 2 gives the UAR results of the Request and Complaint classifiers for the six linguistic feature sets and for two composite sets of linguistic features. Tw+Hw feature set of size 1,536 is the concatenation of Tw and Hw feature sets. Ta+Tc+Ha+Hc feature set of size 3,072 is the concatenation of Ta, Tc, Ha and Hc feature sets.

Table 2: UAR (%) for Linguistic features for Request task

| Feature | Task | Hc | Ha | Hw | Tc | Ta | Tw |
|---|---|---|---|---|---|---|---|
| UAR | Request | 88.2 | 88.9 | **90.8** | 86.6 | 88.2 | 87.8 |
| | Complaint | 65.0 | 67.3 | 66.4 | 65.6 | 64.0 | **68.0** |
| Feature | Task | Tw+Hw | | Ta+Tc+Ha+Hc | | | |
| UAR | Request | **91.2** | | 91.1 | | | |
| | Complaint | 66.5 | | **70.3** | | | |

The best results are most often obtained with the whole textual data (90.8% for the Request task and 68% for the Complaint task) without distinguishing customer text data from agent text data. It seems that the Head truncation is more suitable for the Request task (90.8% versus 87.8%) while the Tail truncation is more suitable for the Complaint task (68% versus 66.4%). It can be noticed that concatenating the feature sets improves the results for Complaint task but not for the Request task.

## 6.2. Vocal Interaction features

Table 3 gives the UAR results of the Request and Complaint classifiers for the two vocal interaction feature sets: Turn Taking (TT) feature set of size 64 and TTc of size 6.

Table 3: UAR (%) for Vocal Interaction features

| Feature | Task | TT | TTc |
|---|---|---|---|
| UAR | Request | 51.9 | 52.4 |
| | Complaint | 54.8 | **57.0** |

The interaction feature sets appear to be suitable for the Complaint task but not at all for the Request task. The TTc feature set confirms its results on the Devel Set by TT feature set.

## 6.3. Audio features

Table 4 gives the UAR results of the Request and Complaint classifiers for the six audio feature sets: three ComPaRe16 audio features set of size 6,373 and three Wav2Vec 2.0 audio features set of size 3,072

Table 4: UAR for Audio features for Request task

| Feat | Cc | Ca | Cj | Wc | Wa | Wj |
|---|---|---|---|---|---|---|
| UAR | 54.1 | 61.3 | 62.2 | 61.9 | 69.4 | **73.7** |
| | 53.0 | 53.5 | 54.8 | 51.6 | 59.5 | **59.9** |

The best results are once more obtained with the joint client-agent channel without distinguishing between the client's and the agent's speech. The Wav2Vec 2.0 features seem more suitable for these two tasks than the ComPaRe16 features. The results are not as good as those obtained by linguistic features.

# 7. Conclusion

In this paper, we have presented the call center Corpus HealthCall recorded in real-life conditions with two separate audio channels, the first one for the customer and the second one for the agent. Each conversation was anonymized respecting the General Data Protection Regulation (GDPR). This corpus includes also a transcription of the spoken conversations.

The first experiment on this database focused on the prediction of CRM annotations from spoken conversions: the types of requests and complaints. For this purpose, we have investigated vocal interaction, audio and linguistic features. The results show that the linguistic features always give the best results (91.2% for the Request task and 70.3% for the Complaint task). The Wav2Vec 2.0 features seem more suitable for these tasks than the ComPaRe16 features (73.7% versus 62.2% for the Request task and 59.9% versus 54.8% for the Complaint task). Vocal interaction features outperformed ComPaRe16 on Complaint task with a 57% rate achieved with only six features.

For future work, we plan to model the temporal evolution of long conversations, to identify different phases such as the identification of the customer, the presentation of the problem and the request for information.